\documentclass[10pt, a4paper, oneside]{article}

\usepackage{amssymb} 
\usepackage{amsmath} 
\usepackage{amsthm} 
\usepackage{hyperref} 
\usepackage{xurl} 
\usepackage{siunitx} 
\usepackage{graphicx} 
\usepackage[labelfont=bf]{caption}
\usepackage{subcaption} 
\usepackage{draftwatermark} 

\newtheorem{theorem}{Theorem}
\newtheorem{lemma}{Theorem}

\begin{document}
\title{Rectifying homographies for stereo vision:\\analytical solution for minimal distortion}

\author{Pasquale Lafiosca$^a$ \and
        Marta Ceccaroni$^b$
}

\date{\small{$^a$Integrated Vehicle Health Management Centre, Cranfield University, United Kingdom} \\
\small{$^b$School of Aerospace, Cranfield University, United Kingdom}
}

\maketitle              

\SetWatermarkText{Article in Press} 
\SetWatermarkScale{3} 

\begin{abstract}
Stereo rectification is the determination of two image transformations (or \textit{homographies}) that map \textit{corresponding points} on the two images, projections of the same point in the 3D space, onto the same horizontal line in the transformed images. Rectification is used to simplify the subsequent \textit{stereo correspondence} problem and speeding up the matching process.
Rectifying transformations, in general, introduce perspective distortion on the obtained images, which shall be minimised to improve the accuracy of the following algorithm dealing with the stereo correspondence problem.
The search for the optimal transformations is usually carried out relying on numerical optimisation. 
This work proposes a closed-form solution for the rectifying homographies that minimise perspective distortion. The experimental comparison confirms its capability to solve the convergence issues of the previous formulation. Its Python implementation is provided.

\end{abstract}


\section{Introduction}\label{secIntro}
Stereo vision gained a prominent role among Computer Vision technologies as it allows machines to perceive depth.
Thanks to pinhole camera model, calibration and epipolar geometry, a considerable simplification can be performed before attempting to solve the stereo correspondence. Among these simplifications, rectification is almost always conducted to obtain horizontal and aligned epipolar lines, so that the following stereo matching algorithm can work along the $x$-axis only, with significant performance increase.

However, rectification transformations generally introduce distortion, which, in turn, impairs the performance of the following stereo matching algorithm. For this reason, minimal-distortion algorithms are proposed in the literature.
In this paper the closed-form solution of the rectifying homographies that minimise perspective distortion is derived by means of the metric introduced by Loop and Zhang~\cite{ComputingRectifyingHomographiesForStereoVision}.
This is applied to a calibrated stereo rig, where all the intrinsic and extrinsic parameters are known.

The hereby presented formulation of the minimising solution is enabled by a new geometrical interpretation of the problem, which simplifies the distortion metric expression.
The solution found is general and, therefore, valid for every relative position of a given stereo rig, even in extreme configurations and when the previous algorithm~\cite{ComputingRectifyingHomographiesForStereoVision} fails to provide an initial guess for minimisation (see Sec.~\ref{sec:Discussion}).
Furthermore, the proposed method overcomes the need for optimisation libraries and is thus computationally more efficient by avoiding numerical minimisation loops.

\subsection{Background}\label{subsecBackground}
Emulating the human vision system, stereo vision derives 3D information by comparison of 2D digital images taken from two distinct camera positions, usually known from calibration~\cite{AFlexibleNewTechniqueForCameraCalibration}. Given a stereo image pair, the third coordinate of each object in the scene is extracted by first solving the \textit{stereo correspondence} problem~\cite{MultipleViewGeometryInComputerVision,HandbookofMachineandComputerVision}, namely the problem of locating the projections of that object in the two images, and then applying \textit{triangulation}~\cite{ActiveOpticalRangeImagingSensors} to recover the position of the object in the 3D world.

Taking advantage of the geometric relations between the 3D points and their projections onto the 2D images, the so-called \textit{epipolar constraint}, the stereo correspondence problem can be reduced to one dimension. This means that the search of each pair of corresponding points is thus carried on along one, usually oblique, \textit{epipolar} line in each picture. To further simplify the problem, \textit{image rectifications} can be applied, transforming the stereo correspondence problem into a search along a horizontal line, with a consequent, significant improvement in efficiency.

Image rectifications are a family of 2D transformations  that can be applied to a couple of non-coplanar stereo images to re-project them onto a \textit{rectifying plane} such that, on the transformed images, corresponding points will lie on the same horizontal line. 
The drawback of rectification is that it introduces a certain amount of distortion in the resulting images, that decreases the accuracy of the subsequent stereo matching algorithms. Hence, the importance of identifying, among the rectifying transformations in the family, the ones that minimise distortion. Hereafter, in continuity with the previous algorithm~\cite{ComputingRectifyingHomographiesForStereoVision}, we will refer to \textit{perspective distortion} simply as distortion.

This paper make the rectifications minimising distortion explicit, finding their analytic formulation via a geometrical derivation. The basic notation and metric are given in Sec.~\ref{sec:Setting}. Our novelties, the geometrical interpretation and derivation of the closed form solution, are reported in Sec.~\ref{sec:Analytical}. The main steps of the algorithm are listed in Sec.~\ref{sec:rectAlgorithmSummary}, followed by a discussion and examples in Sec.~\ref{sec:Discussion}. Conclusions follow in Sec.~\ref{sec:Conclusions}. The Python 3 implementation of the algorithm is also provided. 

\subsection{Previous work}\label{sec:previousWork}
Several algorithms for stereo image rectification are available in the literature. Among the mostly accepted and applied, the one by Fusiello et al.~\cite{ACompactAlgorithmForRectificationOfStereoPairs} essentially fixes the rectifying plane using the sole orientation of one of the two cameras. Despite being a simple and compact, this algorithm is based on an arbitrary non-optimal choice of one rectifying plane (see Sec.~\ref{subsec:rectExamples}). This choice leads, for highly skew configurations of the cameras, to the introduction of a massive amount of distortion in the transformed images, with the consequence of worsening the performances of the algorithm dealing with the stereo correspondence.

The importance of minimising the distortion introduced by the rectifying transformations is highlighted by Loop and Zhang ~\cite{ComputingRectifyingHomographiesForStereoVision}. In their work, the authors decompose rectifying homographies as a combination of a \textit{perspective}, a \textit{similarity} and a \textit{shearing} sub-transformations, and minimise the global distortion. A metric for measuring perspective distortion is thus presented. Finally, using an iterative, numerical procedure, the pair of homographies introducing the minimum perspective distortion is found. However, the initial guess thereby suggested cannot always be found~\cite{ImageRectificationUsingAffineEpipolarGeometricConstraint}.
 
With the aim to minimise distortions at pixel level, a different distortion metric was proposed by Gluckman and Nayar~\cite{RectifyingTransformationsThatMinimizeResamplingEffects}, making use of the Jacobian of the transformation to track changes in local image areas. This method proposes an approximate solution or requires to follow a complex iterative procedure to obtain the optimal one.

A first-attempt of a closed-form solution was proposed by Sun~\cite{ClosedFormStereoImageRectification}. Here the rectification transformations are readily found estimating the fundamental matrix. However, it leads to a particular solution without the criteria of reducing perspective distortion.

Lately the attention of researchers has been focused on strategies for rectifying \textit{uncalibrated} stereo rigs~\cite{ANewImageRectificationAlgorithm,KUMAR20101445,DSRforUncalibratedDualLensCameras,RobustRecoveryOfTheEpipolarGeometryForAnUncalibratedStereoRig,monasse2010three}. These methods do not require a calibration process and try to estimate camera parameters using the scene itself.
For example, a handy solution for uncalibrated dual lens cameras~\cite{DSRforUncalibratedDualLensCameras} relies on key-points that have to be matched in the images. This approach is based on the very limiting assumption of a \textit{small-drift} between the cameras poses. Obviously this does not apply to general camera poses, like verging cameras, and calibration is still important to reach high accuracy and efficiency levels. Different approaches focus on rectifying fisheye images~\cite{Abraham2005FishEyeStereoCA,LearningFisheye,yin2018fisheyerecnet}.
Uncalibrated and fisheye rectification algorithms are outside the scope of this article. To date, the vast majority of stereo vision systems employs calibration, that allows for a \textit{metric} reconstruction with the subsequent triangulation.

\section{Setting the Problem}\label{sec:Setting}

This Section summarises the work of Loop and Zhang~\cite{ComputingRectifyingHomographiesForStereoVision} to provide the basis for the following closed-form solution.

\subsection{Pinhole Cameras and Epipolar Geometry}\label{sub:epi}

\begin{figure}[htp]
 \centering
  \includegraphics[width=\linewidth]{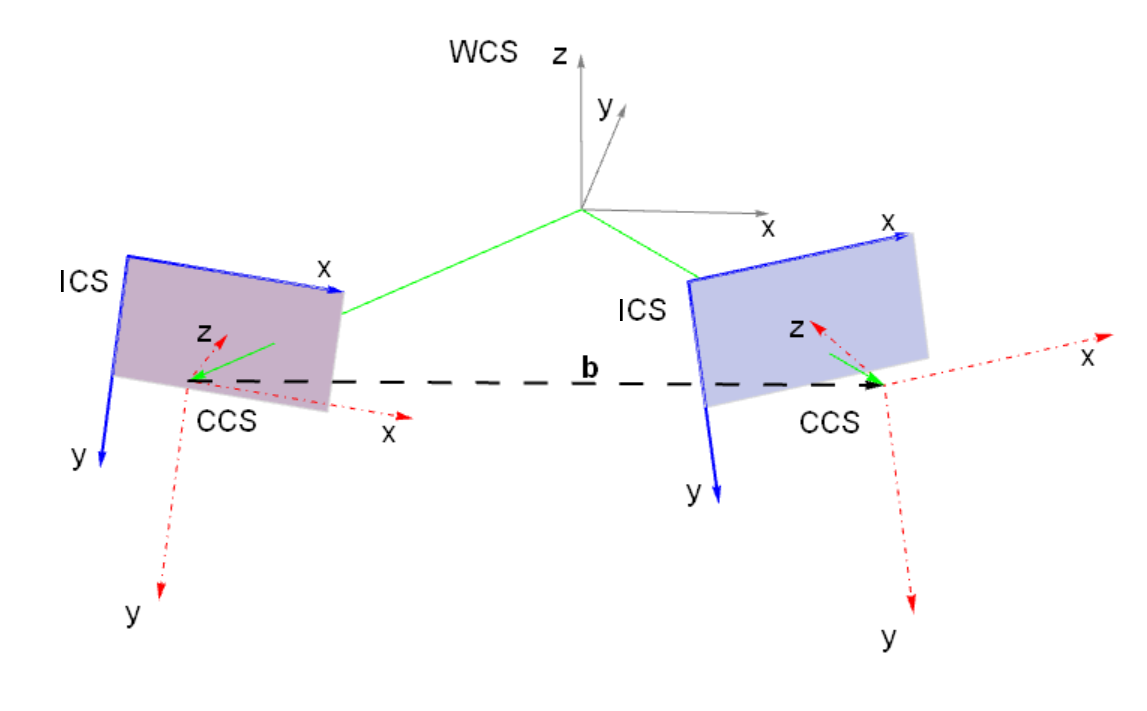}
 \caption{Representation of coordinate systems. CCSs are in red, dot-dashed. ICSs are in blue. WCS is in grey and the baseline is dashed black.}
 \label{fig:coordinateSystems}
\end{figure}

Let us consider a calibrated stereo rig composed of two \textit{pinhole} cameras, where distortion caused by lens imperfections has been already corrected.
As most definitions hereafter will be analogous for both cameras, we will define them only for one camera (definitions for the second camera will be obtained replacing the subscript with 2).
Let $\mathbf{A}_1\in\mathbb{R}^{3\times3}$ be the intrinsic matrix of the left camera, with $\mathbf{R}_1 \in\mathbb{R}^{3\times3}$ and $\mathbf{t}_1 \in \mathbb{R}^{3}$ its rotation matrix and translation vector, respectively, describing the position of the first \textit{Camera Coordinate System} (CCS) with respect to  \textit{World Coordinate System} (WCS), as represented in Fig. \ref{fig:coordinateSystems},  with a slight abuse of notation for axes names to ease visualisation.

Call $\mathbf{o}_1 = -\mathbf{R}_1^{-1} \cdot \mathbf{t}_1$ the position of the optical center in the WCS. Hereafter, unless otherwise stated, elements are expressed in WCS.
The \textit{baseline} is the vector $\mathbf{b} = \mathbf{o}_2 - \mathbf{o}_1$ going from the first to the second camera center.
Additionally, an \textit{Image Coordinate System} (ICS) is defined on the image plane of each camera, where the left image $I_1$ forms (located at $z=1$ in CCS) with the $x$ and $y$ axes parallel to the respective CCS ones and origin in the upper left corner of the image, corresponding to $
\begin{bmatrix} -\frac{[\mathbf{A}_1]_{13}}{[\mathbf{A}_1]_{11}} & -\frac{[\mathbf{A}_1]_{23}}{[\mathbf{A}_1]_{22}} & 1 \end{bmatrix}$ in CCS. The notation $[\mathbf{A}_1]_{ij}$ indicates the $(i,j)$ element of $\mathbf{A}_1$. The ICS system is shown in  Fig. \ref{fig:coordinateSystems} as well.

Given two corresponding image points $\mathbf{p}_1 \in I_1$ and $\mathbf{p}_2 \in I_2$, each one in its ICS and expressed as homogeneous coordinates (i.e. $\mathbf{p}_1 \in \mathbb{R}^3$, with unitary $3^{rd}$ coordinate), the epipolar geometry constraint is defined as:
\begin{equation} \label{eq:epipolarConstraint}
 \mathbf{p}_2^T \cdot \mathbf{F} \cdot \mathbf{p}_1 = 0
\end{equation}
where $\mathbf{F}$ is the fundamental matrix, a $3\times3$ matrix of rank 2 that can only be determined up to a scale factor~\cite{MultipleViewGeometryInComputerVision,TheFundamentalMatrixTheoryAlgorithmsAndStabilityAnalysis}, assumed as known.
$\mathbf{F}$ maps a point on $I_1$ to a \textit{line} on $I_2$, and vice versa (using its transpose).
Given that we work in a \textit{projective space}, all the points are defined up to an arbitrary scaling factor.

On each image all the epipolar lines will intersect in a single point called \textit{epipole}.
Let $\mathbf{e}_1 \in I_1$ and $\mathbf{e}_2 \in I_2$ be the two epipoles in homogeneous ICS, they can be defined as the left and right \textit{kernels} of $\mathbf{F}$, so that:
\begin{equation} \label{eq:epipolesDefinition}
 \mathbf{F} \cdot \mathbf{e}_1 = \begin{bmatrix} 0 & 0 & 0 \end{bmatrix}^T = \mathbf{F}^T \cdot \mathbf{e}_2
\end{equation}
Geometrically, $\mathbf{e}_1$ is the projection of $\mathbf{o}_2$ on the image $I_1$. Similarly for $\mathbf{e}_2$. It is worth noticing that each pair of corresponding epipolar lines lies on a same plane together with the baseline.

\subsection{Rectification}\label{sub:Rectification}
Rectification is the problem of determining two homographies that map corresponding epipolar lines onto parallel horizontal lines sharing the same $y$-coordinate (i.e. sending the epipoles to $\infty$). Thus, rectified images must have the new fundamental matrix in the form~\cite{MultipleViewGeometryInComputerVision}:

\begin{equation} \label{eq:antisymmetricMatrix}
    \overline{\mathbf{F}} = \begin{bmatrix}
    0 & 0 & 0 \\
    0 & 0 & -1 \\
    0 & 1 & 0
    \end{bmatrix}
\end{equation}

Defining the rectified points ${\overline{\mathbf{p}}}_1 = \mathbf{H}_1 \cdot \mathbf{p}_1$ and ${\overline{\mathbf{p}}}_2 = \mathbf{H}_2 \cdot \mathbf{p}_2$, then, from Eq.~(\ref{eq:epipolarConstraint}), we get:
\begin{equation} \label{eq:newFundamentalMatrix}
\begin{split}
 \mathbf{p}_2^T \cdot \mathbf{F} \cdot \mathbf{p}_1 \hspace{0.5em}&= {\overline{\mathbf{p}}}_2^{T} \cdot \mathbf{H}_2^{-T} \cdot \mathbf{F} \cdot \mathbf{H}_1^{-1} \cdot {\overline{\mathbf{p}}}_1 \hspace{0.5em}\\&= {\overline{\mathbf{p}}}_2^{T} \cdot \overline{\mathbf{F}} \cdot {\overline{\mathbf{p}}}_1 \hspace{0.5em}\\&=\hspace{0.5em} 0
\end{split}
\end{equation}
where $\mathbf{H}_1$ and $\mathbf{H}_2$ are the left and right sought-after homographies, respectively.

\subsection{Perspective Distortion}
Following the original procedure~\cite{ComputingRectifyingHomographiesForStereoVision}, given a generic homography $\mathbf{H}_1$, we scale it by its last element and decompose as:
\begin{equation}
 \mathbf{H}_1 = \begin{bmatrix}
    u_{1a} & u_{1b} & u_{1c} \\
    v_{1a} & v_{1b} & v_{1c} \\
    w_{1a} & w_{1b} & 1
    \end{bmatrix} = \mathbf{H}_{1a} \cdot \mathbf{H}_{1p}
\end{equation}
where $\mathbf{H}_{1a}$ is an affine transformation and $\mathbf{H}_{1p}$
is a purely perspective transformation in the form:
\begin{equation}
 \mathbf{H}_{1p} = \begin{bmatrix}
    1 & 0 & 0 \\
    0 & 1 & 0 \\
    w_{1a} & w_{1b} & 1
    \end{bmatrix}
\end{equation}
$\mathbf{H}_{1p}$ is the sole responsible for introducing perspective distortion.
The affine transformation will then be:
\begin{equation}
\mathbf{H}_{1a} = \begin{bmatrix}
    u_{1a}-u_{1c}w_{1a} & u_{1b}-u_{1c}w_{1b} & u_{1c} \\
    v_{1a}-v_{1c}w_{1a} & v_{1b}-v_{1c}w_{1b} & v_{1c} \\
    0 & 0 & 1
    \end{bmatrix}
\end{equation}
$\mathbf{H}_{1a}$ can be decomposed further in a shearing transformation, followed by a similarity transformation~\cite{ComputingRectifyingHomographiesForStereoVision}.
The same decomposition is derived for $\mathbf{H}_{2}$.

Let $\mathbf{p}_{1} = [x_1\ y_1\ 1]^T$ be a generic point on $I_1$, in homogeneous ICS, then the perspective transformed point is:
\begin{equation}
\begin{split}
{\mathbf{p}'}_{1} \hspace{0.5em}&=\hspace{0.5em} \mathbf{H}_{1p} \cdot \mathbf{p}_{1}\\[2pt]
&= \begin{bmatrix} x_1\\
y_1\\
w_{1a}x_1 + w_{1b}y_1 + 1 \end{bmatrix} \\[2pt]
&=\begin{bmatrix} x_1\\
y_1\\
\mathbf{w}_1^T \cdot \mathbf{p}_1 \end{bmatrix}\\[2pt]
&\propto \begin{bmatrix} \frac{x_1}{\mathbf{w}_1^T \cdot \mathbf{p}_1} \\ \frac{y_1}{\mathbf{w}_1^T \cdot \mathbf{p}_1} \\ 1 \end{bmatrix}
\end{split}
\end{equation}
where $\mathbf{w}_1^T = \begin{bmatrix}
w_{1a} &  w_{1b} & 1
\end{bmatrix}^T$.

The codomain of $I_1$ through $\mathbf{H}_{1p}$ must be intended as including the point $\infty$, as per the \textit{hyperplane model} defined to handle perspective transformations~\cite{Gallier2011}.

Notice that if $w_{1a} = w_{1b} = 0$ there is no perspective component in the rectifying transformations, then $\mathbf{H}_1$ is a purely affine transformation. If this is true also for $\mathbf{H}_2$, then the stereo rig is perfectly frontoparallel and the image pair is already rectified. However this, in general, does not happen.

\subsection{Distortion Metric}
We refer to the distortion metric introduced by Loop and Zhang~\cite{ComputingRectifyingHomographiesForStereoVision}. The aim is to select $\mathbf{H}_1$ and $\mathbf{H}_2$ ``as affine as possible'', meaning that the elements $w_{1a}$, $w_{1b}$, $w_{2a}$ and $w_{2b}$ should be chosen so to introduce less distortion.

Taking the average point $\mathbf{p}_{1c} = \frac{1}{n_1}\sum_{i=1}^{n_1} \mathbf{p}_{1i}$ as reference, with $n_1$ total number of pixels of $I_1$, the amount of distortion on $I_1$ is defined as:
\begin{equation}\label{eq:metric}
\sum_{i=1}^{n_1} \frac{ \mathbf{w}_1^T \cdot ( \mathbf{p}_{1i} - \mathbf{p}_{1c}) }{\mathbf{w}_1^T \cdot \mathbf{p}_{1c}}
\end{equation}
The goal is to find the global minimum of the sum of the distortion of both images:
\begin{equation}\label{eq:perspectiveDistortion}
 \sum_{i=1}^{n_1} \frac{ \mathbf{w}_1^T \cdot ( \mathbf{p}_{1i} - \mathbf{p}_{1c}) }{\mathbf{w}_1^T \cdot \mathbf{p}_{1c}}
 +
 \sum_{i=1}^{n_2} \frac{ \mathbf{w}_2^T \cdot ( \mathbf{p}_{2i} - \mathbf{p}_{2c}) }{\mathbf{w}_2^T \cdot \mathbf{p}_{2c}}
\end{equation}
Then (\ref{eq:perspectiveDistortion}) can be rewritten in matrix form as:
\begin{equation}\label{eq:perspectiveDistortionMatrixForm}
 \frac{ \mathbf{w}_1^T \cdot \mathbf{P}_1 \cdot \mathbf{P}_1^T  \cdot \mathbf{w}_1 }{\mathbf{w}_1^T \cdot \mathbf{P}_{c1}\cdot \mathbf{P}_{c1}^T \cdot \mathbf{w}_1}
 +
 \frac{ \mathbf{w}_2^T \cdot \mathbf{P}_2 \cdot \mathbf{P}_2^T  \cdot \mathbf{w}_2 }{\mathbf{w}_2^T \cdot \mathbf{P}_{c2}\cdot \mathbf{P}_{c2}^T \cdot \mathbf{w}_2}
\end{equation}
where:
\begin{equation}
\mathbf{P}_1 \cdot \mathbf{P}_1^T = \frac{w h}{12} \begin{bmatrix}
w^2-1 & 0 & 0 \\
0 & h^2-1 & 0 \\
0 & 0 & 0
\end{bmatrix}
\end{equation}
and:
\begin{equation}
\mathbf{P}_{c1} \cdot \mathbf{P}_{c1}^T = \begin{bmatrix}
\frac{(w-1)^2}{4} & \frac{(w-1)(h-1)}{4} & \frac{(w - 1)}{2} \\
\frac{(w-1)(h-1)}{4} & \frac{(h-1)^2}{4} & \frac{(h - 1)}{2} \\
\frac{(w - 1)}{2} & \frac{(h - 1)}{2} & 1
\end{bmatrix}
\end{equation}
where $w$ and $h$ are pixel width and height of $I_1$. Similarly applies to $I_2$.

\section{Geometric Interpretation and Analytical Derivation of the Minimum}\label{sec:Analytical}
A geometric interpretation of the family of rectifying homographies is introduced hereafter, which paves the way to find the analytical formulation for the global minimum of Eq.~\ref{eq:perspectiveDistortion} and build the corresponding rectifying homographies that minimise perspective distortion.

\subsection{Geometric Interpretation}
\begin{figure}[htp]
 \centering
  \includegraphics[width=\linewidth]{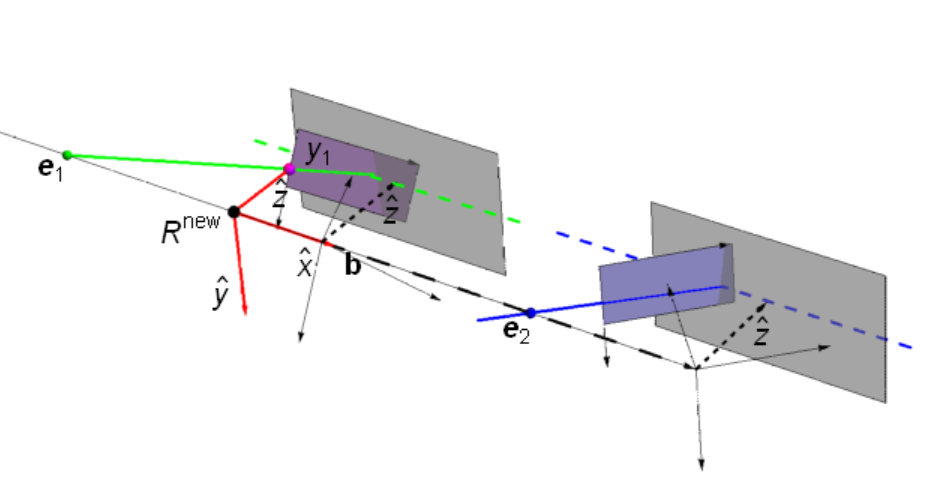}
 \caption{Common orientation of the virtual camera pair (red), projecting on a common plane (gray). $\mathbf{\hat{x}}$ is parallel to the baseline $\mathbf{b}$ (black, dashed). The corresponding epipolar lines (blue, green) and $\mathbf{\hat{z}}$ are identified by $y_1$ (magenta). Rectified epipolar lines are green and blue, dashed.}
 \label{fig:rectifiedCameras}
\end{figure}

It is known that finding rectifying transformations can alternatively be seen as determining a new \textit{common orientation} for a pair of novel \textit{virtual} cameras projecting the images on the same principal plane, such that epipolar lines become horizontal (Fig.~\ref{fig:rectifiedCameras}).

In this subsection we will demonstrate that this can only be achieved if the $\hat{x}$-axis of the new \textit{common orientation} is chosen parallel to the baseline, while the $\hat{z}$-axis can be arbitrarily chosen, thus generating the full rectifying family.

The new common orientation will thus be defined as:
\begin{equation}\label{eq:Rnew}
\mathbf{R}^{new} = \begin{bmatrix}
    \mathbf{\hat{x}}^T \\
    \mathbf{\hat{y}}^T \\
    \mathbf{\hat{z}}^T
    \end{bmatrix}
\end{equation}
where $\mathbf{\hat{x}} = \frac{\mathbf{b}}{ \| \mathbf{b} \| }$ (the versor of the baseline) while $\mathbf{\hat{z}}$ and $\mathbf{\hat{y}}$ can be chosen accordingly to form a Cartesian coordinate system.

Each homography will have to cancel the corresponding intrinsic camera matrix and camera rotation, re-orient the camera with the new chosen orientation and apply another \textit{affine} transformation, namely:
\begin{equation}\label{eq:HShownAsRnew}
    \begin{split}
    \mathbf{H}_{1} = \mathbf{K}_1 \cdot \mathbf{R}^{new} \cdot (\mathbf{A}_1 \cdot \mathbf{R}_1)^{-1} \\
    \mathbf{H}_{2} = \mathbf{K}_2 \cdot \mathbf{R}^{new} \cdot (\mathbf{A}_2 \cdot \mathbf{R}_2)^{-1}
    \end{split}
\end{equation}
where $\mathbf{K}_1, \mathbf{K}_2 \in \mathbb{R}^{3\times3}$ are arbitrary \textit{affine} transformations.

The geometric interpretation above is based on the following Theorem.

\begin{theorem}\label{theorem1}
The epipolar lines of two cameras are corresponding horizontal lines \textit{if and only if} the two cameras share the same orientation, with the $\hat{x}$-axis parallel to their baseline.
\end{theorem}
To prove this theorem we first introduce two Lemmas.

\begin{lemma}\label{lemma1}
The fundamental matrix $\mathbf{F}$, with an abuse of notation, can be written as:
\begin{equation}
\mathbf{F} \propto \mathbf{G} \times \mathcal{H}  = \begin{bmatrix}
\left[\mathbf{G}\right]_2 \left[\mathcal{H}\right]_{3*}- \left[\mathbf{G}\right]_3 \left[\mathcal{H}\right]_{2*}\\
\left[\mathbf{G}\right]_3 \left[\mathcal{H}\right]_{1*}- \left[\mathbf{G}\right]_1 \left[\mathcal{H}\right]_{3*}\\
\left[\mathbf{G}\right]_1 \left[\mathcal{H}\right]_{2*}- \left[\mathbf{G}\right]_2 \left[\mathcal{H}\right]_{1*}\\
\end{bmatrix}
\label{effe}
\end{equation}
where $\mathbf{G}=\mathbf{A}_2\cdot \mathbf{R}_2\cdot\mathbf{b}\in\mathbb{R}^{3}$ and $\mathcal{H} = \mathbf{A}_{2}\cdot \mathbf{R}_{2}\cdot \left(\mathbf{A}_{1}\cdot \mathbf{R}_{1}\right)^{-1}\in\mathbb{R}^{3\times3}$.
The $i^{th}$ row of a matrix is denoted as $ [\  \ ]_{i*}$.
\end{lemma}

\textit{Proof (Lemma~\ref{lemma1})}: let be $\mathbf{X}\in \mathbb{R}^3$ a point in WCS and  $\mathbf{x}_1\in I_1$ and $\mathbf{x}_2\in I_2$ its projections in ICS, then $\mathbf{X}$ is a non-trivial solution of the linear system~\cite{TurriniTrento}: 
\begin{equation}
  \left\{
 \begin{array}{ll}
\mathbf{A}_1\cdot(\mathbf{R}_1\cdot\mathbf{X}+\mathbf{t}_1)= \lambda \mathbf{x_1}\\
\mathbf{A}_2\cdot(\mathbf{R}_2\cdot\mathbf{X}+\mathbf{t}_2)=\mu \mathbf{x_2}
\end{array}
\right.
\end{equation}
where $\lambda, \mu \in \mathbb{R}$. Setting $ \widetilde{\mathbf{X}} = \mathbf{A}_1\cdot(\mathbf{R}_1\cdot\mathbf{X}+\mathbf{t}_1)$, yields:
\begin{equation}
  \left\{
 \begin{array}{ll}
\mathbb{I}_3\cdot\widetilde{\mathbf{X}} &= \lambda \mathbf{x_1}\\
\mathcal{H}\cdot\widetilde{\mathbf{X}}-\mathbf{G} &= \mu \mathbf{x_2}
\end{array}
\right.
 \end{equation}
where $\mathbb{I}_3\in \mathbb{R}^{3\times3}$ is the identity matrix. Therefore: 
  \begin{equation}
  Det\left(
  \left[
  \begin{array}{cccc}
  \mathbb{I}_3&\mathbf{0}& \mathbf{x}_1&\mathbf{0}\\
 \mathcal{H}&\mathbf{G} & \mathbf{0}&\mathbf{x}_2\\
  \end{array}
  \right]
  \right)=0
 \end{equation}
which is linear in both $\mathbf{x}_1$ and $\mathbf{x}_2$ and can thus be rewritten as:
 \begin{equation}
  \mathbf{x}_2^T\cdot \mathbf{F}\cdot \mathbf{x}_1 = 0
  \end{equation}
with:
\begin{equation}  
  [\mathbf{F}]_{ij}=(-1)^{i+j}Det(\hat{\mathcal{Q}}^{i,j}),\hspace{2em} \forall i,j=1,2,3.
 \end{equation}
where $\hat{\mathcal{Q}}^{i,j}\in \mathbb{R}^{4\times4}$ is equal to the matrix
$
\mathcal{Q}=\left[
  \begin{array}{cc}
  \mathbb{I}_3&\mathbf{0} \\
 \mathcal{H}&\mathbf{G} \\
  \end{array}
  \right] \in \mathbb{R}^{6\times6}$ with the $i^{th}$ and $(3+j)^{th}$ rows dropped.
Furthermore:
\begin{equation}
 [\mathbf{F}]_{ij} = (-1)^{i+1}Det(\bar{\mathcal{S}}^{i,j})
 \label{dimostrazione}
\end{equation}
with the first column of $\bar{\mathcal{S}}^{i,j}\in \mathbb{R}^{2\times2}$ equal to the $j^{th}$  column of the matrix $\mathcal{H}$ with the $i^{th}$ row dropped, and the second column equal to $\mathbf{G}$ with the $i^{th}$ element dropped. Then Eq.~(\ref{dimostrazione}) is equivalent to Eq.~(\ref{effe}) $\square$.

\textit{Remark:} it can be shown that the matrix $\mathcal{H}$ is the matrix representing the axes of the first CCS as seen from the second CCS
and that $ \mathbf{e}_2=\frac{\mathbf{G}}{\|\mathbf{G}\|}$.
Therefore the fundamental matrix can be expressed as
$
\mathbf{F}\propto \mathbf{e}_2 \times \mathcal{H}
$, similarly to Eq.~(\ref{effe}).

\begin{lemma}\label{lemma2}
The epipolar lines of the second camera are horizontal \textit{if and only if} the $x$-axis of the second camera is parallel to the baseline.
\end{lemma}

\emph{Proof (Lemma~\ref{lemma2}):} by the definition of the fundamental matrix (see Sec \ref{sub:epi}), the epipolar lines of the second camera are horizontal \textit{if and only if} the first row of the fundamental matrix $\mathbf{F}$ is null.
Using Eq.~\textit{Lemma~\ref{lemma1}} the first row of $\mathbf{F}$ can be expressed as:
\begin{equation}
\left[\mathbf{G}\right]_2 \left[\mathcal{H}\right]_{3*}- \left[\mathbf{G}\right]_3 \left[\mathcal{H}\right]_{2*}
 \label{qui}
\end{equation}
From the remark above, the rows of $\mathcal{H}$ form a base of linearly independent vectors, therefore the only linear combination that can set the formula~(\ref{qui}) to zero 
is the trivial one, with  $ \left[\mathbf{G}\right]_2 = \left[\mathbf{G}\right]_3=0$, that is equivalent to state that the $x$-axis of the second camera is parallel to the baseline $\square$.

We can now demonstrate \textit{Theorem~\ref{theorem1}}.

\emph{Proof (Theorem~\ref{theorem1}):} by \textit{Lemma~\ref{lemma1}} and \textit{Lemma~\ref{lemma2}}, if the epipolar lines of the second camera are horizontal, the fundamental matrix can be reduced to the form:
 \begin{equation}
 \overline{\mathbf{F}}\propto\begin{bmatrix}
\mathbf{0}\\
- \left[\mathcal{H}\right]_{3*}\\
\hspace{1em}\left[ \mathcal{H}\right]_{2*}\\
\end{bmatrix}
 \end{equation}
Using Eq.~(\ref{eq:antisymmetricMatrix}) we impose:
  \begin{equation}
  \begin{split}
    \left[\mathcal{H}\right]_{2*} &= \begin{bmatrix}0 & 1 & 0\end{bmatrix}\\
    \left[\mathcal{H}\right]_{3*} &= \begin{bmatrix}0 & 0 & 1\end{bmatrix}
  \end{split}
 \end{equation}

That is equivalent to require that the $y$ and $z$ axes of the first CCS, as seen from the second CCS system of reference (respectively $ \left[\mathcal{H}\right]_{2*}$ and  $\left[\mathcal{H}\right]_{3*}$), must be parallel to the corresponding axes of the second camera $\square$.

\subsection{Analytic Derivation of the Minimising Rectification}
By Eqs.~(\ref{eq:Rnew}) and (\ref{eq:HShownAsRnew}), finding the global minimum of Eq.~(\ref{eq:perspectiveDistortion}) is equivalent to find the optimal common orientation $\mathbf{R}^{new}$ of the virtual cameras.
Since $\mathbf{\hat{x}}$ is imposed by the baseline, the problem is reduced to finding the $\mathbf{\hat{z}}$ so that distortion is minimised. Setting $\mathbf{\hat{y}} = \mathbf{\hat{z}} \times \mathbf{\hat{x}}$, will then determine $\mathbf{R}^{new}$.

Choosing the minimising $\mathbf{\hat{z}}$, in turn, is equivalent, by construction, to finding the pair of corresponding epipolar lines to become the new \textit{horizon lines} (see Fig.~\ref{fig:rectifiedCameras}), which will lay at the intersection of the plane $\mathbf{\hat{y}}=0$ and the rectified images. Actually only one \textit{horizon line} must be determined, as the corresponding one is readily found by means of $\mathbf{F}$.

Finally, the minimisation problem is reduced to a single parameter problem, noticing that an epipolar line is uniquely determined by its $y$-intercept $(0,y_1) \in I_1$ in ICS. Such $y$-intercept, in WCS, takes the form: 
\begin{equation}
\mathbf{y}_1 = \mathbf{R}_1^{-1} \cdot (\mathbf{A}_1^{-1} \cdot \begin{bmatrix}0 & y_1 & 1\end{bmatrix}^T - \mathbf{t}_1)
\end{equation}

The $\mathbf{\hat{z}}$  axes of the virtual cameras will thus be the direction of the line $\mathbf{z}$  perpendicular to $\mathbf{b}$, passing through $\mathbf{y}_1$, calculated as the difference between the vector going from $\mathbf{o}_1$ to $\mathbf{y}_1$ and its projection on the baseline (its derivation is done by means of the outer product $\otimes$~\cite{MultilinearSubspaceLearning}):
\begin{equation}\label{eq:zeta}
\begin{split}
\mathbf{z} &= ( \mathbf{y}_1 - \mathbf{o}_1 ) - [( \mathbf{y}_1 - \mathbf{o}_1 )^T \cdot \mathbf{\hat{x}} ] \mathbf{\hat{x}}\\ 
&= ( \mathbf{y}_1 - \mathbf{o}_1 ) - \mathbf{\hat{x}} \otimes \mathbf{\hat{x}} \cdot ( \mathbf{y}_1 - \mathbf{o}_1 )
\end{split}
\end{equation}
Then follows:
\begin{equation}
    \mathbf{\hat{z}} = \frac{ \mathbf{z} }{ \lVert \mathbf{z} \rVert }
\end{equation}
Using Eqs.~(\ref{eq:Rnew}), (\ref{eq:HShownAsRnew}) and (\ref{eq:zeta}), yields:
\begin{equation}
    \begin{split}
    \mathbf{w}_1^T &= [ \mathbf{H}_1 ]_{3*} \\
                 &= [ \mathbf{K}_1 \cdot \mathbf{R}^{new} \cdot (\mathbf{A}_1 \cdot \mathbf{R}_1)^{-1}]_{3*}  \\
                 &= \mathbf{\hat{z}}^T \cdot (\mathbf{A}_1 \cdot \mathbf{R}_1)^{-1} \\
                 &= \frac{ \mathbf{z}^T }{ \| \mathbf{z} \| } \cdot (\mathbf{A}_1 \cdot \mathbf{R}_1)^{-1} \\
                 &= \left( \mathbf{R}_1^{-1} \cdot \mathbf{A}_1^{-1} \cdot \begin{bmatrix} 0 \\ y_1 \\ 1 \end{bmatrix} - \mathbf{\hat{x}} \otimes \mathbf{\hat{x}} \cdot \mathbf{R}_1^{-1} \cdot \mathbf{A}_1^{-1} \cdot \begin{bmatrix} 0 \\ y_1 \\ 1 \end{bmatrix} \right)^T\\
                 &\hspace{1.5em} \cdot (\mathbf{A}_1 \cdot \mathbf{R}_1)^{-1} \\
                 &= \begin{bmatrix} 0 & y_1 & 1 \end{bmatrix} \cdot (\mathbf{A}_1 \cdot \mathbf{R}_1)^{-T} \cdot (\mathbb{I}_3 - \mathbf{\hat{x}} \otimes \mathbf{\hat{x}})\\
                 &\hspace{1.5em}\cdot (\mathbf{A}_1 \cdot \mathbf{R}_1)^{-1} \\
    \end{split}
\end{equation}
where $\mathbf{K}_1$ has been omitted as it does not affect the $3^{rd}$ row of what follows (being an affine transformation, its last row is $\begin{bmatrix}
0 & 0 & 1
\end{bmatrix}$), and
$\lVert \mathbf{z} \rVert$ has been discarded noticing that in Eq.~(\ref{eq:perspectiveDistortionMatrixForm}) it appears both at numerator and denominator.
Rearranging:
\begin{equation}
\begin{split}
 \mathbf{w}_1 &= (\mathbf{A}_1 \cdot \mathbf{R}_1)^{-T} \cdot (\mathbb{I}_3 - \mathbf{\hat{x}} \otimes \mathbf{\hat{x}}) \cdot (\mathbf{A}_1 \cdot \mathbf{R}_1)^{-1} \cdot \begin{bmatrix} 0 \\ y_1 \\ 1 \end{bmatrix}\\
 &= \mathbf{L}_1 \cdot \begin{bmatrix} 0 \\ y_1 \\ 1 \end{bmatrix}
\end{split}
\end{equation}
where the matrix:
\begin{equation}
\mathbf{L}_1 =  (\mathbf{A}_1 \cdot \mathbf{R}_1)^{-T} \cdot (\mathbb{I}_3 - \mathbf{\hat{x}} \otimes \mathbf{\hat{x}}) \cdot (\mathbf{A}_1 \cdot \mathbf{R}_1)^{-1}
\end{equation}
Similarly for $\mathbf{w}_2$:
\begin{equation}
 \mathbf{w}_2 = \mathbf{L}_2 \cdot \begin{bmatrix} 0 \\ y_1 \\ 1 \end{bmatrix}
\end{equation}
where:
\begin{equation}
\mathbf{L}_2 =  (\mathbf{A}_2 \cdot \mathbf{R}_2)^{-T} \cdot (\mathbb{I}_3 - \mathbf{\hat{x}} \otimes \mathbf{\hat{x}}) \cdot (\mathbf{A}_1 \cdot \mathbf{R}_1)^{-1}
\end{equation}

Our goal is to minimise the distortion from Eq.~(\ref{eq:perspectiveDistortionMatrixForm}), that can now be written as:
\begin{equation}
\begin{split}
 \frac{ \begin{bmatrix} 0 & y_1 & 1 \end{bmatrix} \cdot \mathbf{M}_1 \cdot \begin{bmatrix} 0 & y_1 & 1 \end{bmatrix}^T }{\begin{bmatrix} 0 & y_1 & 1 \end{bmatrix} \cdot \mathbf{C}_1 \cdot \begin{bmatrix} 0 & y_1 & 1 \end{bmatrix}^T}\\[2pt]
 +\frac{ \begin{bmatrix} 0 & y_1 & 1 \end{bmatrix} \cdot \mathbf{M}_2 \cdot \begin{bmatrix} 0 & y_1 & 1 \end{bmatrix}^T }{\begin{bmatrix} 0 & y_1 & 1 \end{bmatrix} \cdot \mathbf{C}_2 \cdot \begin{bmatrix} 0 & y_1 & 1 \end{bmatrix}^T}\label{eq:perspectiveDistortionNewForm1}
\end{split}
\end{equation}
where:
\begin{equation}
 \begin{split}
  \mathbf{M}_1 &= \mathbf{L}_1^T \cdot \mathbf{P}_1 \cdot \mathbf{P}_1^T \cdot \mathbf{L}_1 \\
  \mathbf{M}_2 &= \mathbf{L}_2^T \cdot \mathbf{P}_2 \cdot \mathbf{P}_2^T \cdot \mathbf{L}_2 \\
  \mathbf{C}_1 &= \mathbf{L}_1^T \cdot \mathbf{P}_{c1} \cdot \mathbf{P}_{c1}^T \cdot \mathbf{L}_1 \\
  \mathbf{C}_2 &= \mathbf{L}_2^T \cdot \mathbf{P}_{c2} \cdot \mathbf{P}_{c2}^T \cdot \mathbf{L}_2
 \end{split}
\end{equation}

The terms of Eq.~(\ref{eq:perspectiveDistortionNewForm1}) can be now expanded as polynomial expressions in $y_1$.
\begin{figure}[htb]
 \centering
  \includegraphics[width=0.9\linewidth]{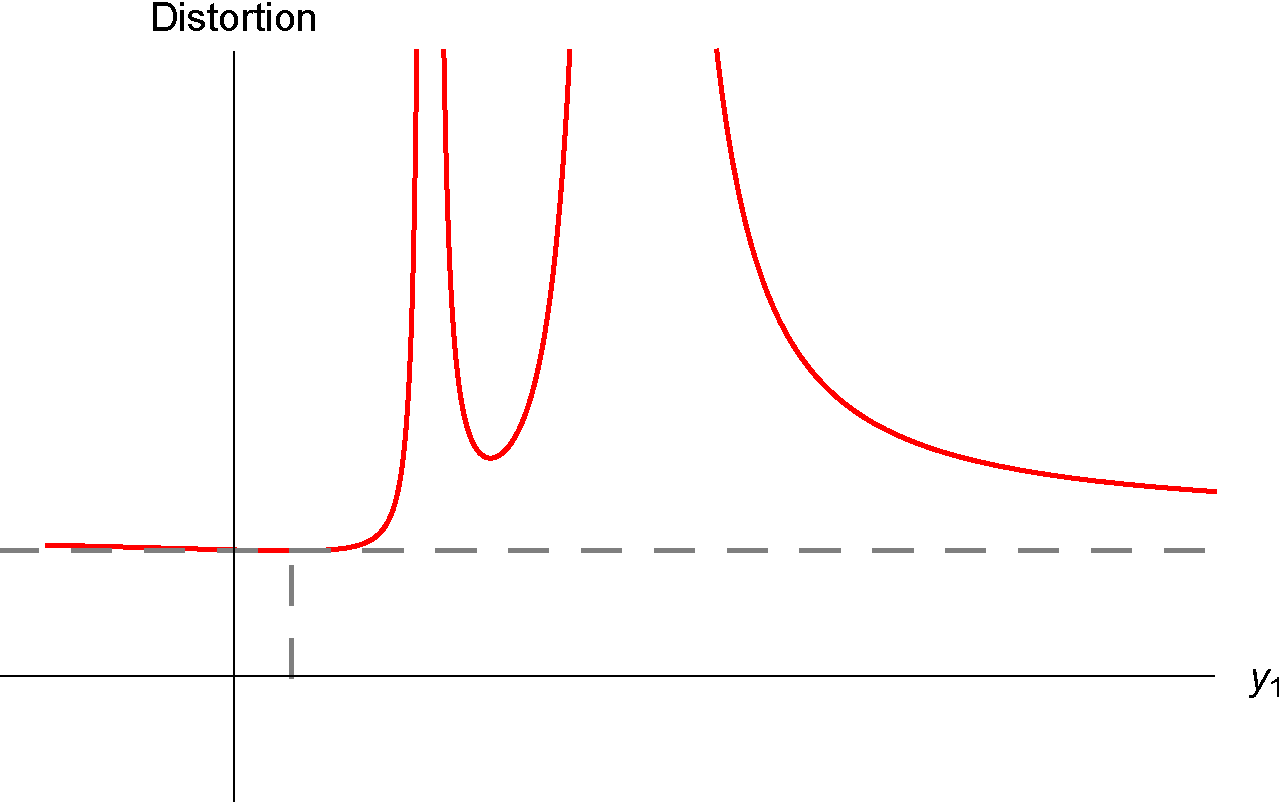}
 \caption{Possible trend of the distortion in Eq.~(\ref{eq:perspectiveDistortionNewForm1}) as a function of $y_1$. Global minimum is identified by dashed lines.}
 \label{fig:Distortion}
\end{figure}
It can be verified that Eq.~(\ref{eq:perspectiveDistortionNewForm1}) takes the form:
\begin{equation}
\frac{f_1(y_1)}{\left(g_1(y_1)\right)^2}+\frac{f_2(y_1)}{\left(g_2(y_1)\right)^2}
\label{eq:fg}
\end{equation}
with $f_1 = f_1(y_1)$, $f_2 = f_2(y_1)$ second degree polynomials and $g_1 = g_1(y_1)$, $g_2 = g_2(y_1)$ first degree polynomials.
Deriving to find the extreme points yields:
\begin{equation}
\left(\frac{f_1}{g_1^2}+\frac{f_2}{g_2^2}\right)'=\frac{g_1 g_2^3 f'_1 + g_1^3 g_2f'_2 - 
 2 f_1 g_2^3 g'_1 - 
 2 f_2 g_1^3 g'_2 }{g_1^3g_2^3}
\end{equation}
where the terms of $5^{th}$ degree cancel out.
Discarding the denominator\footnote{It can be shown that approaching the roots of the denominator of the distortion function both from the upper and lower limit, the function always goes to $+\infty$, therefore the global minimum never reaches $-\infty$, as in Fig.~\ref{fig:Distortion}.}, the extreme points are thus found as the solution of a $4^{th}$ degree polynomial:
\begin{equation}\label{eq:4thdegree}
 a y_1^4 + b y_1^3 + c y_1^2 + d y_1 + e = 0
\end{equation}
with $a, b, c, d, e \in \mathbb{R}$.
The solutions of Eq.~(\ref{eq:4thdegree}) can be found using any solving formula for homogeneous polynomials of $4^{th}$ degree. Full calculations are reported in Appendix A.

Fig.~\ref{fig:Distortion} shows a possible behavior of Eq.~(\ref{eq:perspectiveDistortionNewForm1}).
Among the acceptable real solutions of Eq.~(\ref{eq:4thdegree}), the one representing the global minimum depends on the initial position of the cameras, and therefore can only be determined by directly comparing the value of the distortion in Eq.~(\ref{eq:perspectiveDistortionNewForm1}) for each solution.
 
It must be remarked that, in the very peculiar (unrealistic) case in which the two cameras are identical (i.e. $\mathbf{A}_1 = \mathbf{A}_2$, $\mathbf{P}_1 = \mathbf{P}_2$ and $\mathbf{P}_{c1} = \mathbf{P}_{c2}$) and share the exact same orientation $\mathbf{R}_1 = \mathbf{R}_2$ (not necessarily frontoparallel), then~(\ref{eq:fg}) takes the form:
\begin{equation}
2\frac{f(y_1)}{\left(g(y_1)\right)^2}
\end{equation}
leading, once derived, to a first degree polynomial, thus to a single minimum.

The analytical expression of $\mathbf{w}_1$ and $\mathbf{w}_2$ minimising perspective distortion is therefore found, so that $\mathbf{H}_{1p}$ and $\mathbf{H}_{2p}$ are directly determined.
In order to find the complete expression of the rectifying homographies $\mathbf{H}_1$ and $\mathbf{H}_2$, the respective affine components, as shearing and similarity transformations, can be easily calculated~\cite{ComputingRectifyingHomographiesForStereoVision}.

\section{Algorithm Summary}\label{sec:rectAlgorithmSummary}
The steps of the direct rectifying algorithm explained above are summarised as follows:
\begin{enumerate}
 \item The weights $\mathbf{w}_1$ and $\mathbf{w}_2$ of the perspective components of the rectifying homographies are written as polynomial expression in $y_1$, the $y$-intercept of the \textit{horizon} epipolar line on $I_1$.
 \item Distortion is written as function of $y_1$ and auxiliary matrices are defined to calculate the coefficients of the quartic polynomial.
 \item The acceptable solutions (either $2$ or $4$) are calculated.
 \item The global minimum is determined by direct comparison of the distortion values obtained.
 \item The minimising $\mathbf{w}_1$ and $\mathbf{w}_2$ are calculated as last row of the matrices in Eq.~(\ref{eq:HShownAsRnew}).
 \item The similarity and shearing transformations are calculated~\cite{ComputingRectifyingHomographiesForStereoVision}.
 \item The rectification transformations $\mathbf{H}_1$ and $\mathbf{H}_2$ are fully determined.
\end{enumerate}
\begin{sloppypar}
The full Python 3 code is made available at \url{https://github.com/decadenza/DirectStereoRectification}.
\end{sloppypar}

\section{Discussion}\label{sec:Discussion}

Rectifying homographies are transformations that dramatically reduce the search space for correspondences between a couple of stereo images, thus making the stereo matching problem computationally affordable.

The optimal transformations are traditionally found making use of numerical minimisation  (see~\cite{ComputingRectifyingHomographiesForStereoVision}), which prevents the use of rectification when low computational power is available (e.g. space assets). Furthermore, the application of the traditional method is limited by a convergence issue, due to the matrices decomposition used therein and thus not avoidable, that will be discussed in the reminder of this Section. For both these reasons, most present applications~\cite{xia2019accurate,hinzmann2018mapping} prefer to use the arbitrary, non-optimal solution proposed by Fusiello et al.~\cite{ACompactAlgorithmForRectificationOfStereoPairs}, that, for peculiar configurations introduces high distortion levels (Sec.~\ref{subsec:rectExamples}), thus impairing performance of the following stereo matching algorithm.

This paper explicits and demonstrates the formula for the optimal rectifying homograpies. The formula is valid for every pair of stereo images, independently from the configuration parameters. Being an exact formulation, it eliminates the need for minimisation, while still providing the optimal transformation, thus enabling the use of rectification in scenarios with very limited computational capabilities, as for example for autonomous navigation of space satellites. Future work will concentrate on such applications.

The reminder of this section discusses a limiting convergence issue of the traditional method proposed by Loop and Zhang~\cite{ComputingRectifyingHomographiesForStereoVision}. Indeed, the convergence of this algorithm cannot be guaranteed as it depends upon finding a suitable initial guess, derived assuming $\mathbf{A}$ and $\mathbf{A'}$ to be positive-definite, which cannot be guarenteed for all configurations~\cite{ImageRectificationUsingAffineEpipolarGeometricConstraint}.

Consider, for example, the case in which the first CCS is coincident with the WCS and an identical second camera is placed in $\begin{bmatrix} 1 & a & b\end{bmatrix}^T$ (with $a, b \in\mathbb{R}$), oriented as the first one but rotated around the $x$-axis of an angle $\theta_x$.
For all cases in which $b = a \tan{\theta_x}$, the first element of $\mathbf{A'}$ will be null, thus, as a straightforward consequence of \textit{Sylvester's Theorem}, $\mathbf{A'}$ will not be positive-definite. Then, Loop and Zhang's algorithm fails.

Moreover, numerical simulations confirm that the configurations for which positive-definite assumption is violated are a numerous and therefore relevant in limiting the applicability of the traditional algorithm.
Indeed, considering a setup with fixed intrinsic and randomly generating one million of extrinsic parameters (i.e. relative position and rotation between the two cameras), it was found that Loop and Zhang's algorithm failed in over the 50\% of the cases, in spite of all the configurations being perfectly legitimate.
This happened mostly because of numerical errors in computing Cholesky decomposition of $\mathbf{A}$ and $\mathbf{A'}$. The proposed analytic method, instead, does not need an initial guess and directly provides the optimal solution.

To the best of the authors's knowledge, for the cases mentioned above, all algorithms present in the the literature fail in providing the optimal rectifying homography (introducing minimum distortion), found instead by the formula derived in this work and not affected by the configuration settings.

\begin{figure}[ht]
\centering
  \begin{subfigure}[t]{\linewidth}
    \includegraphics[width=\linewidth]{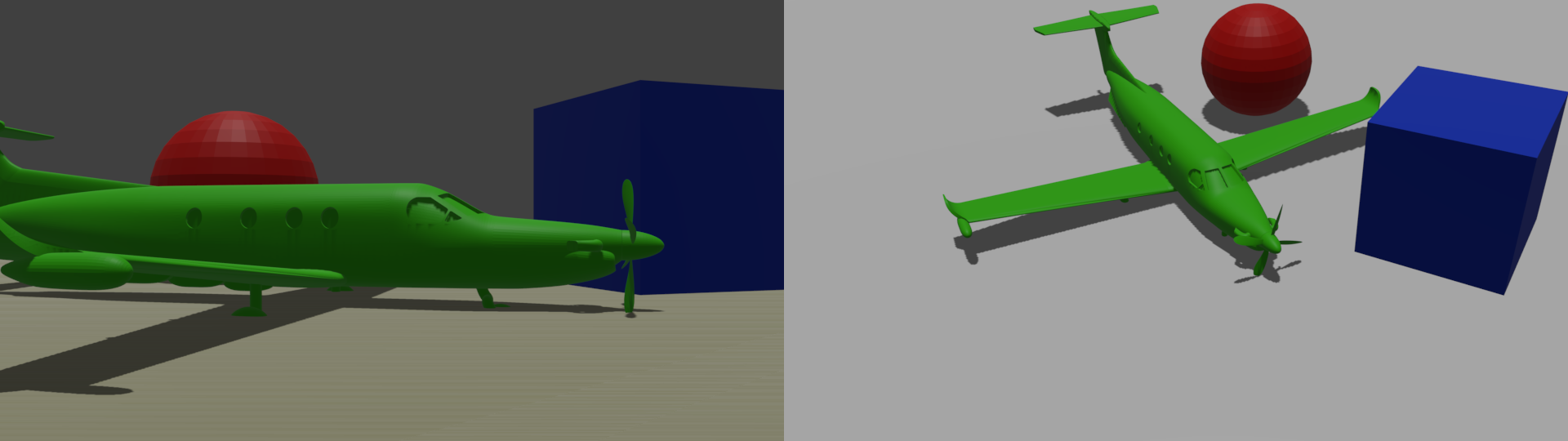} 
  \end{subfigure}
  \par\medskip
  \begin{subfigure}[b]{\linewidth}
    \includegraphics[width=\linewidth]{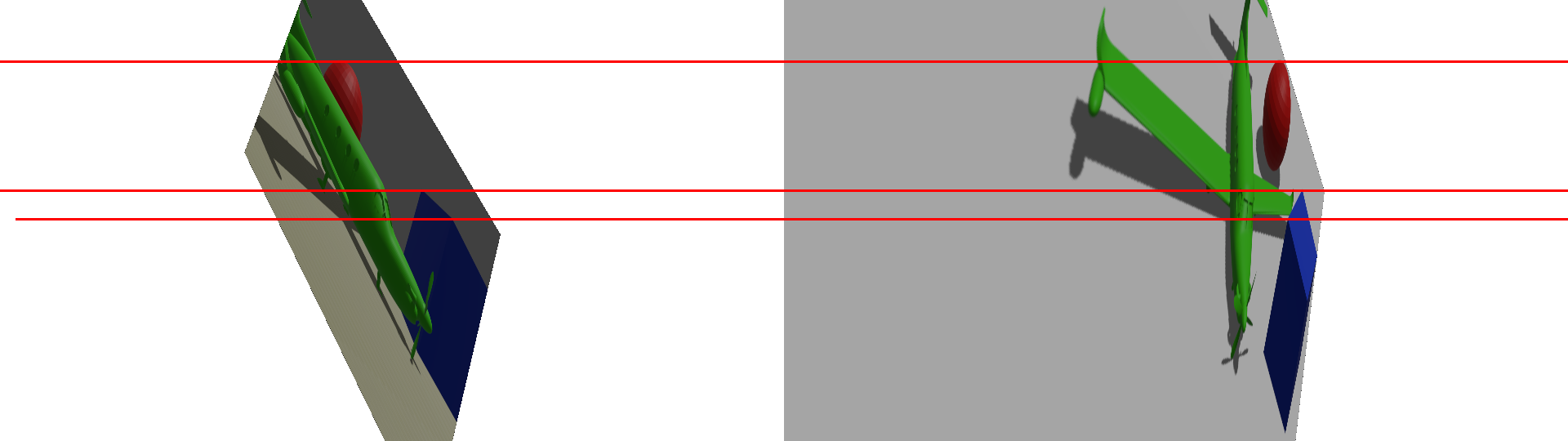} 
  \end{subfigure}

  \caption{Left and right image of a synthetic scene with extremely skewed camera positions (top-left and top-right, respectively). In this case the numerical minimisation of~\cite{ComputingRectifyingHomographiesForStereoVision} fails. Corresponding rectified image pair using our algorithm is showed (bottom). Horizontal lines are drawn for reference on the rectified image.}
  \label{fig:exampleExtreme}
\end{figure}

Furthermore, Fig.~\ref{fig:exampleExtreme} shows the optimal solution found by our algorithm in case of extremely skewed camera positions configuration, for which Loop and Zhang's initial guess cannot be retrieved.
Finally it must be mentioned that the direct algorithm here presented is capable of providing the optimal solution also in the case of one camera entering the field of view of the other, i.e. when the epipoles lie within image boundaries, which causes convergence issues when employing numerical algorithms.

In summary, the formula here presented guarantees convergence for every configuration, avoids the need of external libraries and saves computational time as it does not make use of minimisation.
Indeed, comparing the computational cost, in our implementation on an Intel i7-7500U CPU 2.70GHz, the original algorithm by Loop and Zhang requires $\SI{8.899}{\milli\second}$ on average to calculate the rectifying homographies, while the proposed method takes $\SI{7.655}{\milli\second}$.

\subsection{Rectification Examples}\label{subsec:rectExamples}

\begin{figure}[ht]
\centering
  \begin{subfigure}[t]{\linewidth}
    \includegraphics[width=\linewidth]{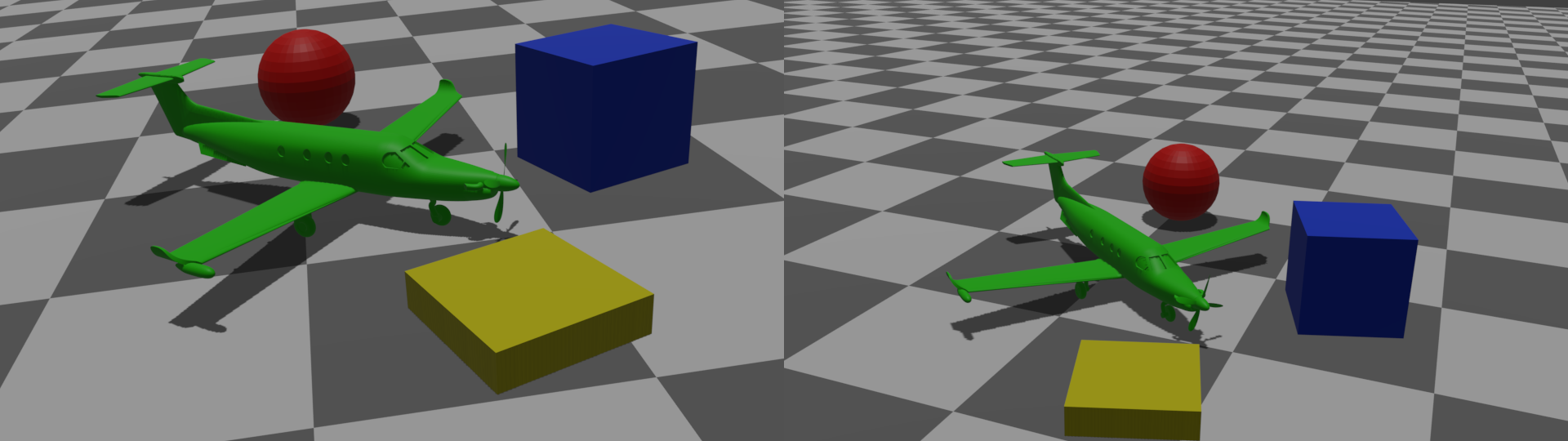} 
  \end{subfigure}
  \par\medskip
  \begin{subfigure}[b]{\linewidth}
    \includegraphics[width=\linewidth]{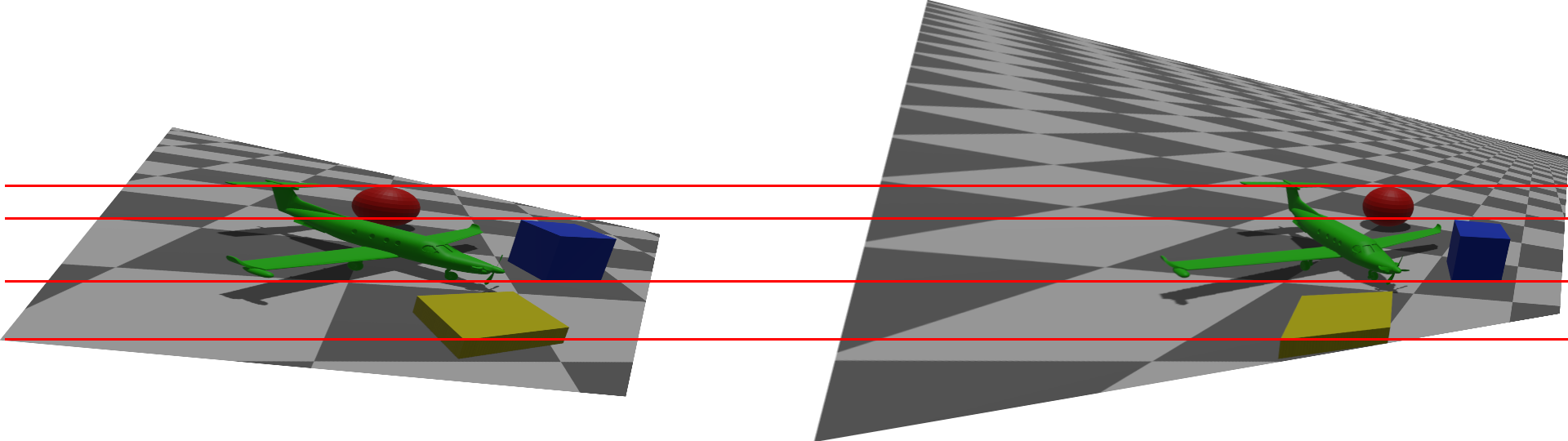}
  \end{subfigure}
 \par\medskip
  \begin{subfigure}[b]{\linewidth}
    \includegraphics[width=\linewidth]{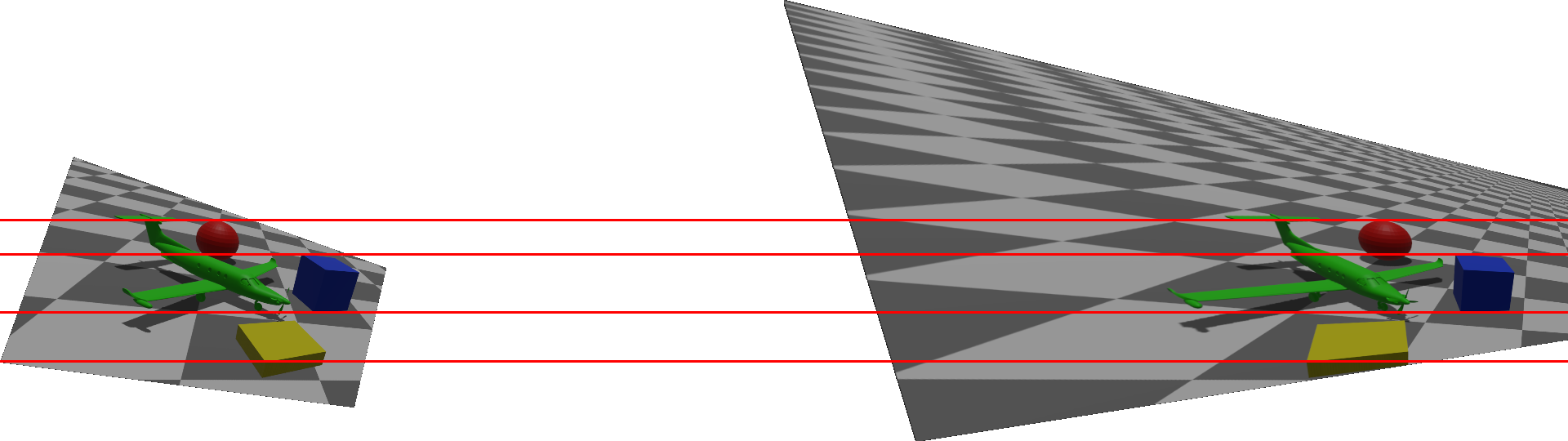} 
  \end{subfigure}
  \caption{Left and right image of a synthetic scene (top-left and top-right, respectively) and corresponding rectified image pair as rectified using the direct algorithm (center) and the algorithm in~\cite{ACompactAlgorithmForRectificationOfStereoPairs}(bottom). Horizontal lines are drawn for reference on the rectified images.}
  \label{fig:example}
\end{figure}

A synthetic example of rectification is shown in Fig.~\ref{fig:example}. The original left and right images are first shown. The images, as rectified following our direct algorithm, are then listed, where arbitrary horizontal lines have been drawn as a reference. Here the effects of rectification are clearly visible, as corresponding points are aligned. This example is included in the Python code.

The couple of images on the third line of Fig.~\ref{fig:example}, shows the same image pair rectified following the algorithm in~\cite{ACompactAlgorithmForRectificationOfStereoPairs}. As expected, while the value of the minimum distortion introduced in the second row is $\SI{46252}{}$, the homographies found by Fusiello's algorithm and generating the third pair of images are non optimal and introduce a distortion of $\SI{48207}{}$, about $4\%$ higher than the minimum.

\begin{figure*}[!ht]
\centering
  \begin{subfigure}[t]{\linewidth}
    \includegraphics[width=\linewidth]{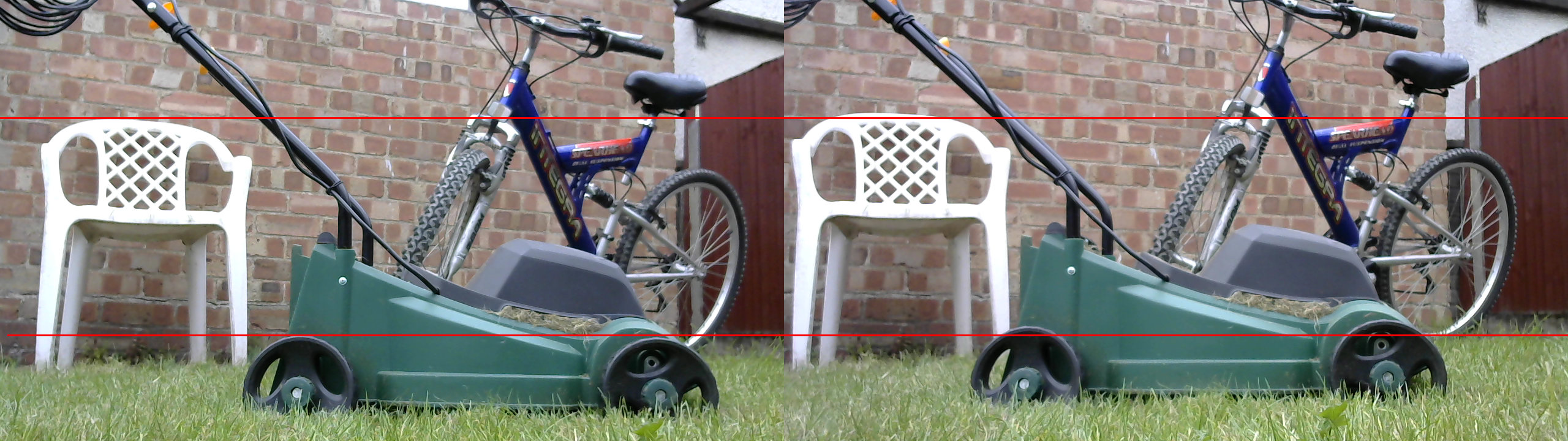} 
  \end{subfigure}
  \par\medskip
  \begin{subfigure}[b]{\linewidth}
    \includegraphics[width=\linewidth]{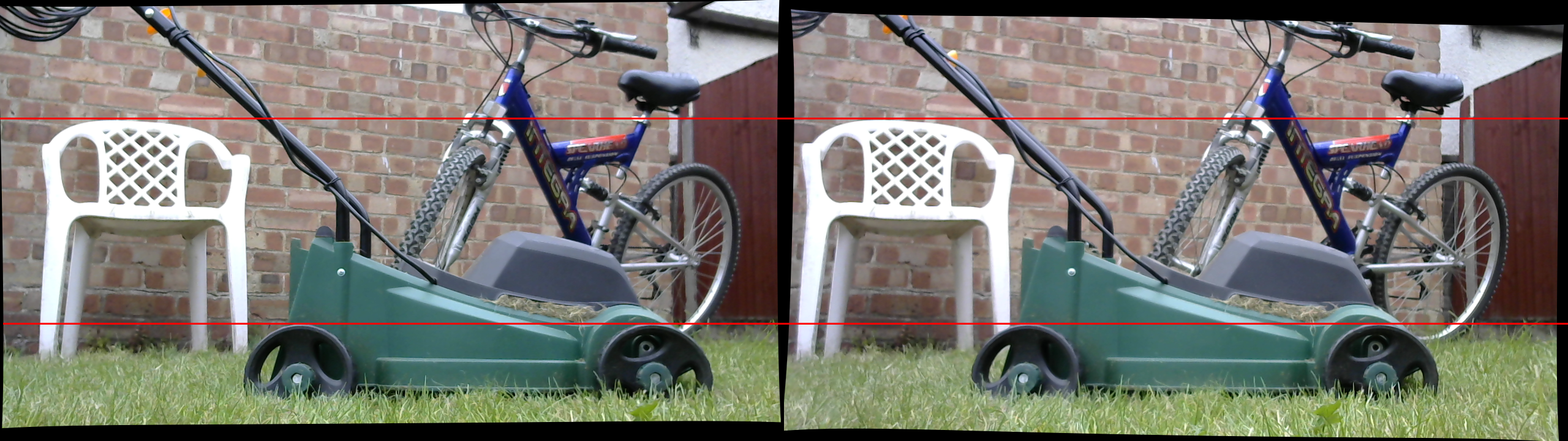} 
  \end{subfigure}
  \caption{Left and right image of a real scene (top-left and top-right, respectively) and corresponding rectified image pair (bottom). Horizontal lines are drawn for reference on both images, showing effects of rectification. Lens distortion correction is also noticeable.}
  \label{fig:exampleReal}
\end{figure*}

In Fig.~\ref{fig:exampleReal} a real scene stereo pair is rectified. Despite the cameras are placed almost frontoparallel, corresponding points still lie on different scan-lines, so rectification is needed.
In real applications, calibration~\cite{AFlexibleNewTechniqueForCameraCalibration} is required to accurately fit the stereo rig to the pinhole camera model.
Apart from lens distortion correction, the algorithm finds no difference between real-case and synthetic images, since it starts from the same premises (i.e. both camera projection matrices).

\section{Conclusions}\label{sec:Conclusions}
A direct rectification algorithm for calibrated stereo rigs has been proposed. Our method improves the well-known approach by Loop and Zhang to calculate the optimal rectifying homographies.

Thanks to an alternative geometrical interpretation of the problem, the proposed algorithm is able to find the formula for the rectifying homographies that introduce the minimal perspective distortion for any camera configurations, even in extremely skewed relative positions, and without depending on minimisation libraries.

\begin{sloppypar}
The Python 3 code has been made publicly available at \url{https://github.com/decadenza/DirectStereoRectification}.
\end{sloppypar}

Because of the lower computational cost, the value of having an analytic solution may be particularly relevant for hardware-limited applications (e.g. space applications, miniaturised devices), where each change of camera extrinsic or intrinsic parameters would require the computation of new rectifying homographies.
Future work might formulate the analytic solution for a distortion metric that includes pixel resampling effects and applications to scenarios with very limited computational capabilities.

\bibliographystyle{ieeetr}
\bibliography{BIBLIOGRAPHY}

\begin{thebibliography}{10}

\bibitem{ComputingRectifyingHomographiesForStereoVision}
C.~Loop and Z.~Zhang, ``Computing rectifying homographies for stereo vision,''
  in {\em Computer Vision and Pattern Recognition, 1999. IEEE Computer Society
  Conference on.}, vol.~1, 1999.

\bibitem{AFlexibleNewTechniqueForCameraCalibration}
Z.~Zhang, ``A flexible new technique for camera calibration,'' {\em Pattern
  Analysis and Machine Intelligence, IEEE Transactions on}, 2000.

\bibitem{MultipleViewGeometryInComputerVision}
R.~Hartley and A.~Zisserman, {\em Multiple View Geometry in Computer Vision}.
\newblock Cambridge, UK: Cambridge University Press, 2~ed., 2003.

\bibitem{HandbookofMachineandComputerVision}
Hornberg, {\em Handbook of Machine and Computer Vision.}
\newblock Wiley‐VCH, 2017.

\bibitem{ActiveOpticalRangeImagingSensors}
P.~Besl, {\em Active Optical Range Imaging Sensors.}
\newblock New York, NY: Springer, 1989.

\bibitem{ACompactAlgorithmForRectificationOfStereoPairs}
A.~Fusiello, E.~Trucco, and A.~Verri, ``A compact algorithm for rectification
  of stereo pairs,'' {\em Machine Vision and Applications}, 2000.

\bibitem{ImageRectificationUsingAffineEpipolarGeometricConstraint}
L.~Sui, Z.~Jiulong, and C.~Duwu, ``Image rectification using affine epipolar
  geometric constraint,'' 2009.

\bibitem{RectifyingTransformationsThatMinimizeResamplingEffects}
J.~Gluckman and S.~Nayar, ``Rectifying transformations that minimize resampling
  effects,'' in {\em Proceedings of the 2001 IEEE Computer Society Conference
  on Computer Vision and Pattern Recognition. CVPR 2001}, 2001.

\bibitem{ClosedFormStereoImageRectification}
C.~Sun, ``Closed-form stereo image rectification,'' Association for Computing
  Machinery, 2012.

\bibitem{ANewImageRectificationAlgorithm}
Z.~Chen, C.~Wu, and H.~Tsui, ``A new image rectification algorithm,'' {\em
  Pattern Recognition Letters}, 01 2003.

\bibitem{KUMAR20101445}
S.~Kumar, C.~Micheloni, C.~Piciarelli, and G.~Foresti, ``Stereo rectification
  of uncalibrated and heterogeneous images,'' {\em Pattern Recognition
  Letters}, 2010.

\bibitem{DSRforUncalibratedDualLensCameras}
X.~Ruichao, S.~Wenxiu, P.~Jiahao, Y.~Qiong, and R.~Jimmy, ``Dsr: Direct
  self-rectification for uncalibrated dual-lens cameras,'' {\em International
  Conference on 3D Vision}, 2018.

\bibitem{RobustRecoveryOfTheEpipolarGeometryForAnUncalibratedStereoRig}
R.~Deriche, Z.~Zhang, Q.-T. Luong, and O.~Faugeras, ``Robust recovery of the
  epipolar geometry for an uncalibrated stereo rig,'' {\em Proceedings of the
  Third European Conference on Computer Vision, vol. 1, Stockholm, Sweden},
  2001.

\bibitem{monasse2010three}
P.~Monasse, J.-M. Morel, and Z.~Tang, ``Three-step image rectification,'' in
  {\em BMVC 2010-British Machine Vision Conference}, 2010.

\bibitem{Abraham2005FishEyeStereoCA}
S.~Abraham and W.~F{\"o}rstner, ``Fish-eye-stereo calibration and epipolar
  rectification,'' {\em Isprs Journal of Photogrammetry and Remote Sensing},
  2005.

\bibitem{LearningFisheye}
X.~Zhucun, N.~Xue, G.-S. Xia, and W.~Shen, ``Learning to calibrate straight
  lines for fisheye image rectification,'' 04 2019.

\bibitem{yin2018fisheyerecnet}
X.~Yin, X.~Wang, J.~Yu, M.~Zhang, P.~Fua, and D.~Tao, ``Fisheyerecnet: A
  multi-context collaborative deep network for fisheye image rectification,''
  in {\em Proceedings of the European conference on computer vision (ECCV)},
  2018.

\bibitem{TheFundamentalMatrixTheoryAlgorithmsAndStabilityAnalysis}
Q.-T. Luong and O.~Faugeras, ``The fundamental matrix: Theory, algorithms, and
  stability analysis,'' {\em International Journal of Computer Vision}, 1996.

\bibitem{Gallier2011}
J.~Gallier, {\em Basics of Projective Geometry}.
\newblock Springer New York, 2011.

\bibitem{TurriniTrento}
C.~Turrini, ``Geometria per la ricostruzione tridimensionale,'' 2017.

\bibitem{MultilinearSubspaceLearning}
H.~Lu, K.~Plataniotis, and A.~Venetsanopoulos, {\em Multilinear Subspace
  Learning: Dimensionality Reduction of Multidimensional Data}.
\newblock Chapman \& Hall/CRC, 01 2013.

\bibitem{xia2019accurate}
R.~Xia, R.~Su, J.~Zhao, Y.~Chen, S.~Fu, L.~Tao, and Z.~Xia, ``An accurate and
  robust method for the measurement of circular holes based on binocular
  vision,'' {\em Measurement Science and Technology}, 2019.

\bibitem{hinzmann2018mapping}
T.~Hinzmann, J.~L. Sch{\"o}nberger, M.~Pollefeys, and R.~Siegwart, ``Mapping on
  the fly: Real-time 3d dense reconstruction, digital surface map and
  incremental orthomosaic generation for unmanned aerial vehicles,'' in {\em
  Field and Service Robotics}, 2018.

\end{thebibliography}

\appendix
\section*{Appendix A}\label{AppendixA}
The coefficients of the $4^{th}$ order polynomial expression in $y_1$ of Eq.~(\ref{eq:4thdegree}) are:
\begin{equation}
\begin{split}
a &= m_2 m_4+m_6m_8\\
b &= m_1 m_4+ 3 m_2 m_3  m_4+ m_5 m_8+3 m_6 m_7 m_8\\
c &= 3  m_1 m_3 m_4+3 m_2 m_3^2 m_4+3 m_5 m_7 m_8+3 m_6 m_7^2 m_8\\
d &= 3 m_1 m_3^2 m_4+ m_2 m_3^3 m_4+3 m_5 m_7^2 m_8+ m_6 m_7^3 m_8\\
e &= m_1 m_3^3 m_4+ m_5 m_7^3 m_8
\end{split}
\end{equation}
with:
\begin{equation}
\begin{split}
m_1 &= \left[\mathbf{M}_1\right]_{2 3} \left[\mathbf{C}_1\right]_{2 3}-\left[\mathbf{M}_1\right]_{3 3} \left[\mathbf{C}_1\right]_{2 2}\\[2pt]
m_2 &= \left[\mathbf{M}_1\right]_{2 2} \left[\mathbf{C}_1\right]_{2 3}-\left[\mathbf{M}_1\right]_{2 3} \left[\mathbf{C}_1\right]_{2 2}\\[2pt]
m_3 &= \frac{\left[\mathbf{C}_2\right]_{2 3}}{\left[\mathbf{C}_2\right]_{2 2}}\\[2pt]
m_4 &= \frac{\left[\mathbf{C}_2\right]_{2 2}}{\left[\mathbf{C}_1\right]_{2 2}}\\[2pt]
m_5 &= \left[\mathbf{M}_2\right]_{2 3} \left[\mathbf{C}_2\right]_{2 3}-\left[\mathbf{M}_2\right]_{3 3} \left[\mathbf{C}_2\right]_{2 2}\\[2pt]
m_6 &= \left[\mathbf{M}_2\right]_{2 2} \left[\mathbf{C}_2\right]_{2 3}-\left[\mathbf{M}_2\right]_{2 3} \left[\mathbf{C}_2\right]_{2 2}\\[2pt]
m_7 &= \frac{\left[\mathbf{C}_1\right]_{2 3}}{\left[\mathbf{C}_1\right]_{2 2}} \\[2pt]
m_8 &= \frac{\left[\mathbf{C}_1\right]_{2 2}}{\left[\mathbf{C}_2\right]_{2 2}}
\end{split}
\end{equation}

The four roots of the equation are given by:
\begin{equation}
y_1=\frac{-b}{4a}\pm Q\pm \frac{1}{2}\sqrt{-4 Q^2-2p+\frac{S}{Q}}
\end{equation}
with:
\begin{equation}
    \begin{split}
    Q &= \frac{1}{2}\sqrt{-\frac{2}{3}p+\frac{1}{3a}\left(\Delta_0+\frac{q}{\Delta_0}\right)}\\
    S &= \frac{8 a^2d-4 a b c +b^3}{8 a^3}\\
    \Delta_0 &= \left(\frac{s+\sqrt{s^2-4q^3}}{2}\right)^{\frac{1}{3}}\\
    p &= \frac{8 a c -3 b^2}{8 a^2} \\
    q &= 12 a e -3 b d +c^2\\
    s &= 27 a d^2-72 a c e + 27 b^2 e -9 b c d +2 c^3
    \end{split}
\end{equation}

\textit{Remark}: for the case $\mathbf{A}_1 = \mathbf{A}_2$, $\mathbf{P}_1 = \mathbf{P}_2$, $\mathbf{P}_{c1} = \mathbf{P}_{c2}$ and $\mathbf{R}_1 = \mathbf{R}_2$, the solution is given by:
\begin{equation}
y_1=-\frac{m_1}{m_2}
\end{equation}

\end{document}